# Toward Corpus Size Requirements for Training and Evaluating Depression Risk Models Using Spoken Language

*Tomek Rutowski, Amir Harati, Elizabeth Shriberg, Yang Lu, Piotr Chlebek, Ricardo Oliveira*

Ellipsis Health

`{tomek, amir, liz, yang, piotr, ricardo}@ellipsishealth.com`

## Abstract

Mental health risk prediction is a growing field in the speech community, but many studies are based on small corpora. This study illustrates how variations in test and train set sizes impact performance in a controlled study. Using a corpus of over 65K labeled data points, results from a fully crossed design of different train/test size combinations are provided. Two model types are included: one based on language and the other on speech acoustics. Both use methods current in this domain. An age-mismatched test set was also included. Results show that (1) test sizes below 1K samples gave noisy results, even for larger training set sizes; (2) training set sizes of at least 2K were needed for stable results; (3) NLP and acoustic models behaved similarly with train/test size variations, and (4) the mismatched test set showed the same patterns as the matched test set. Additional factors are discussed, including label priors, model strength and pre-training, unique speakers, and data lengths. While no single study can specify exact size requirements, results demonstrate the need for appropriately sized train and test sets for future studies of mental health risk prediction from speech and language.

## 1. Introduction

Depression is a rising global health concern associated with high levels of burden on individuals, families and society [1][2]. To help fill gaps in patient care, research has explored whether technology can be used for remote care management tasks including patient screening and monitoring. Spoken language has been of particular interest because it is natural, requires only a personal device, and offers cues from both language and acoustic-prosodic information [3][4][5][6].

Privacy constraints on health data present a challenge for obtaining and sharing large labeled data sets; this is particularly true for spoken (as opposed to written) language. As a result, while valuable progress has been made through research and common corpora [7][8][9][10][11][12], most past studies on depression and spoken language are based on small amounts of data. The number of labeled sessions typically ranges from several dozen [8][13][14][15] to several hundred [16][17]. The number of unique speakers generally ranges from below 50 [18] to 600 [4]. Recording lengths per label range from a few seconds to half an hour [8]. In some cases, datasets are too small for machine learning, and research describes only feature associations [16][17][18].

Results of these studies are often hard to replicate on other datasets [19]. Many studies use either a single separate subset of matched data, or have used a cross-validation approach for classification [13][20][21] and regression models [8][20][21] have also been investigated. Performance metrics include F1, AUC, accuracy, precision, recall and sensitivity, specificity, MAE, and RMSE [13][19][20][21]. Studies often use test and development sets that are matched to the training data in collection characteristics. Given the limited data available, it is not unusual to obtain large differences between development and test results [8][22].

In clinical applications, it is critical to estimate model robustness. Depression risk prediction from spoken language offers promise, but corpus size limits in research make it difficult to estimate performance in the wild. Variability around test set performance needs to be better understood. This study provides results on performance variability as a function of both test and training set size. Results use a large corpus that supports a range of sample sizes for both train and test sets. While it is not feasible to provide absolute guidelines for studies differing in corpora, models, and other factors, the goal is to share a set of reference results in this domain, and to illustrate the need for larger data sets in future work.

The metric used is AUC for binary classification; data set sizes too small to support this metric will also be too small for larger class counts or for regression. The study includes two model types: one based on information from natural language processing (NLP model) and the other based on information from the speech signal (acoustic model). Both utilize current deep learning approaches in the literature to provide a general idea of how such models perform. The study also includes results for both matched versus mismatched evaluation data.

## 2. Method

### 2.1. Data

The current study requires large amounts of depression-labeled spoken language from which to sample a wide range of train and test set sizes. A large number of speakers is needed to allow training, development and test partitions with no speaker overlap, and to represent speaker variation in samples of different sizes. Despite the availability of common corpora for depression prediction from spoken language [7][8][9], no common datasets meeting these criteria were available. For this reason, the study utilizes American English data from Ellipsis Health. The data cannot currently be shared due to privacy restrictions; Ellipsis Health is exploring ways to collect less restricted data in the future. Note that the goal of the paper is to share information on general data size effects rather than to benchmark performance. It is believed that for the former purpose, the use of proprietary data is appropriate.

#### 2.1.1. "General population" (GP) corpus

The primary corpus in the study is referred to as the "general population" corpus. (The term is not meant to imply overarching demographic coverage). Table 1 shows information on corpus partitions. There is no speaker overlap across partitions. Speakers in this set ranged in age from 18 to 65 with a mean age of 33. Participants were recruited from all 50 states in the U.S. Speakers were paid to interact for roughly five minutes with a software application on their personal device. The application posed personal questions on different topics such as "concerns" and "home life". Users answered by speaking freely in response to the questions. Speakers initiated the start and end points of their responses in the application. The average number of responses per session was 4.6. Responses in this collection averaged about 80 seconds, which corresponds to roughly 180 words in length. The sum of train, development and test responses is 65K.

Table 1. *Statistics by partition for GP and SP corpora.*

| Partition | GP-Train | GP-Dev | GP-Test | SP-Test |
|---|---|---|---|---|
| # Responses | 51,543 | 7,392 | 7,072 | 7,018 |
| # Speakers | 6,579 | 1,578 | 1,499 | 264 |
| # Sessions | 11,606 | 1,578 | 1,499 | 1,304 |
| % Depressed | 28.41 | 20.96 | 22.36 | 30.27 |
| US States | 50 | 50 | 50 | 1 |
| Avg. Age | 33 | 32 | 32 | 64 |
| Resp./ Session | 4.4 | 4.7 | 4.7 | 6.0 |
| %Female | 59.4 | 59.6 | 58.7 | 59.0 |

*2.1.2. "Senior population" (SP) corpus*

A second Ellipsis Health corpus was included to provide an example of train/test mismatch in the study. The corpus will be referred to as the Senior Population (SP) corpus because speakers came from a retirement community. Statistics are shown in Table 1. Patients in this collection used an application similar to that described for the GP data. Speakers ranged in age from 45 to 75 (99% of data), with a mean age of 64. In contrast to the GP data, participants came from a specific community in California, and met on site with study administrators to receive instructions. Unlike the GP collection, participants in the SP corpus signed on to participate for six sessions. They were compensated based on their actual participation. Patients in this study used different devices than were used in GP collection. The average number of responses per session was 6.0; responses averaged 35 seconds and 76 words in length.

*2.1.3. Depression risk class labels*

For both the GP and SP corpora, class labels were obtained from the PHQ-8, a version of the Patient Health Questionnaire (PHQ-9) [23] after the suicidality question was removed. Participants completed the PHQ-8 in the application, after providing their spoken responses. PHQ-8 outcomes were then thresholded for binary classification. Following a clinical standard [24][25], scores at or above 10 were mapped to positive for depression risk and scores below 10 were mapped to negative for depression risk.

**2.2. Study design**

*2.2.1. Train and test sizes*

The unit of data was the response. Training data was sampled from the GP-train subset (51K responses). Five training sizes were defined: **200, 500, 2K, 20K, and 51K.** (In retrospect, adding training sizes between 2K and 20K is recommended). Test data was sampled from either the GP-test (7K) or the SP-test (7K) data. Six test sizes were defined: **200, 500, 1K, 2K, 5K and 7K**. Crossing the train with the test sizes results in 30 conditions. As described earlier, most past work has used data set sizes near or below 500. By including much larger set sizes, the study aims to discover how additional data reduces performance variability.

*2.2.2. Sampling within train/test size pairs*

For each train/test size pair (for example 500 train and 1K test), 25 experiments were run. The 25 experiments are the result of crossing five random samplings for the training size with five random samplings for the test size. Random sampling was done with replacement. The purpose of the 25 experiments is to estimate the variability of results given a fixed model and a fixed training/test size pair. The number 25 was chosen to balance estimation and feasibility.

*2.2.3. Development data (fixed size)*

For all experiments, a fixed development set of 7K responses was used. The reason for using a fixed development set was that there were over 2,000 unique experiments to run. The reason for using a large development size was to avoid biasing results using a possibly poor fixed random sampling for each of the test sizes. A consequence of using this large development set is that variability in performance in the study is *underestimated*—particularly for the smaller data sizes as also noted in Figure 1.

**2.3. Models**

*2.3.1. NLP Model*

The goal of including the NLP model is to help predict trends for NLP models generally in this domain. For the NLP task, the speech signal was first transcribed using a publicly available ASR service. Word error rate was roughly 20%. From internal experiments, it was observed that high ASR error rates were tolerated by the NLP model; this is likely based on good cue redundancy. The NLP model is based on a transformer architecture and takes advantage of transfer learning from a language modeling task [26]. A DeBERTa [27] pre-trained model is used. It was chosen because it outperforms RoBERTA [28], ALBERT [29], BERT [30], in our experiments.

DeBERTa has the advantage of having fewer (435M) parameters relative to other comparable models, while still providing good performance. This was useful given the large number of experiments to run. The DeBERTa model used in this paper is pre-trained on over 80GB of text data from the following common corpora; Wiki, Books and OpenWebtext. The input context window was 512 long and the tokenizer was trained for 128K tokens.

For fine-tuning, a predictor head was attached to the language model and a binary classifier was trained. For all experiments, all hyperparameters other than the learning rate were fixed. The learning rate was set proportionally to the amount of training data for each experiment (grid search approach). Early stopping was used to avoid extensive runtime utilization. Additional information on earlier versions of this model can be found in [31] and [32].

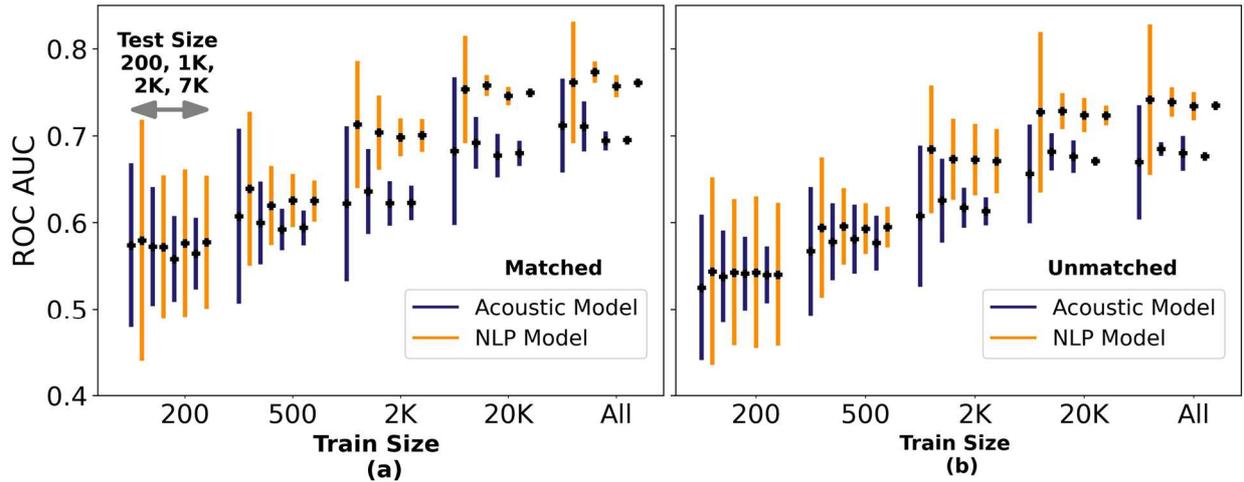

Figure 1- *(a): Variation in AUC by condition. Each vertical bar displays the mean and two standard deviations above and below the mean for a given test size. Vertical bars for test sizes of 500 and 5K were similar to those for 200 and 7K respectively and are removed for lack of space. 1- (b): Same as (a) but testing on SP data (train/test mismatch). All experiments use a fixed large development set; therefore, bar sizes are underestimates of true variation.*

### 2.3.2. Acoustic Model

The goal of including an acoustic model is to represent how acoustic and signal-based models may behave generally in this domain. The acoustic model used is based on an encoder-decoder architecture. Models using other deep learning architectures, including long-short term memory (LSTM) and convolutional neural networks (CNNs) performed less well in separate experiments. The acoustic model works in two stages. Speech is first segmented every 25 seconds. The model learns a latent representation at the segment level, using filter-bank coefficients as input. Representations are then fused to make a prediction at the response or session level. Model training uses transfer learning from an automatic speech recognition (ASR) task. The ASR decoder is discarded after the pre-training stage; the encoder is fine-tuned along with the predictor layer.

The encoder consists of a CNN followed by layers of LSTM. The predictor layer uses a Recurrent CNN (RCNN) [33]. The last layer of the RCNN module is used as a vector representation of a given audio segment. It is passed along with other representations of the same session to a fusion module that makes a final prediction for the session. The fusion model uses a max operation on all vectors to obtain an aggregate representation for the session, and then uses a multilayer perceptron to make a final prediction. Information on earlier versions of this model can be found in [32] and [34].

## 3. Results and Discussion

### 3.1. Effect of test set size

Results of the experiments are shown in Figure 1. Figure 1-(a) depicts results for training and testing on the GP data. Figure 1-(b) shows results for training on the GP data and testing on the SP (i.e. unmatched) data. Each vertical bar displays the mean AUC and 2 standard deviations above and below the mean for 25 randomly drawn train/test sets using the indicated train and test sizes. For example, the first bar on the left of Figure 1-(a) shows results for the acoustic model, GP train size 200, GP test set size 200, using 25 samplings for those set sizes. Of the 6 test sets evaluated (200, 500, 1K, 2K, 5K and 7K), only 4 (200, 1K, 2K, 7K) are shown due to space constraints. The results for test size 500 were close to those for set size 200. Results for test size 5K were close to those for test size 7K. All experiments use a fixed, large 7K development set to limit the number of conditions and to focus on comparisons of test sizes without worrying about development sets being too small to be reliable. The variability shown is thus likely to be an underestimate of variability in experiments conducted with development sets smaller than 7K.

### 3.2. Effect of train set size

For training sizes of 200 and 500, both models show low mean AUC and high variance. The models differ in amount of training data needed to see an appreciable increase in mean AUC. Whereas the NLP model shows a mean AUC increase at 2K, the acoustic model shows a mean AUC increase at the next shown step, or 20K. This difference may reflect differences in model pre-training. The NLP model uses a transformer-based model pre-trained on 80GB of text data. The acoustic model is pre-trained on 1K hours of speech.

Note that the amount of required test data decreases as the amount of training data increases. For models trained on only 2K data points, a test set of at least 7K data points was required to obtain stable results. An overall suggestion based on these observations pertains to estimating the value of additional data. One method is to use trends based on performance variability from past sets, and assess the gain from new data. The recommendation here is to evaluate not based on only a single obtained performance value, but rather on the variability in performance observed when sampling multiple times from the added data.

### 3.3. Example with train/test mismatch

To provide an example of results in the case of a mismatch between training and test data, Figure 1(b) shows the same experimental conditions as 1(a), but replacing GP test data with SP test data. There was no model tuning or retraining for the SP data. As would be expected, overall performance is lower than for matched test data. Nevertheless, the pattern of variation in AUC with train and test sizes is remarkably similar to that for the matched data.

Table 2. *Test sizes needed for a 5% relative AUC variation. NA means there is no set large enough to satisfy the requirement.*

| Model | Train-200 | Train-500 | Train-2K | Train-20K | Train-50K |
|---|---|---|---|---|---|
| Acoustic | NA | NA | NA | 5,000 | 2,000 |
| NLP | NA | NA | 7,000 | 1,000 | 1,000 |

Table 3. *Test sizes needed for a 20% relative AUC variation. NA means there is no set large enough to satisfy the requirement.*

| Model | Train-200 | Train-500 | Train-2K | Train-20K | Train-50K |
|---|---|---|---|---|---|
| Acoustic | 2,000 | 500 | 500 | 500 | 200 |
| NLP | NA | 500 | 500 | 200 | 200 |

### 3.4. Significance testing for training set size

To test for significant differences in the mean AUC across set sizes in the study, an unpaired two-sided unequal-variance Welch's t-test was used to compare the 25 AUC results in one train/test size to the 25 AUC results of the next adjacent train/test size. Due to space limitations, results for variation in training size are reported while fixing test set size to 200, 500, 1K, 2K, and 5K. For each test set size, the following comparisons for adjacent steps in the experiment training sizes were compared: 200 to 500, 500 to 2K, and 2K to 20K. This resulted in 3 training-size comparisons for each of the 5 different test sizes. All of the resulting 15 t-tests yielded significant differences at $p <= 0.001$, showing that a step up in training data size produced a significant increase in AUC.

### 3.5. Variability tolerance and chance performance

Table 2 and Table 3 provide examples of the amount of data needed in this study for two different tolerances of AUC variability around the mean. Consider a mean AUC of 0.70; a 5% tolerance gives a range from 0.675 to 0.735 whereas 20% gives a range from 0.63 to 0.77. To compare experiments close in performance it seems more useful to use a tolerance closer to 5%. Table 2 shows the minimum test size for NLP and acoustic models as training size changes to obtain AUC variations of less than 5% around the mean (2.5% above and below). For small training sizes there is no test size (up to 7K in the experiments) that results in a less than 5% relative variation. Table 3 shows results for 20% tolerance.

Figure 1 suggests that studies using data sets with sizes below 500 data points may have performance distributions that overlap values associated with chance performance. The minimum training size for NLP that results in AUC above 0.60 is 500 data points. However, if the goal is to reach 5% tolerance, there is no available test size in this study that is large enough (NA in Tables 2 and 3). For the acoustic model the minimum training size is 2K however, there is no test size available that results in 5% relative AUC variation. Although the NLP model shows better-than-chance performance with 2K data points for training, it requires 7K data points for a relatively stable result. Therefore 2K data points seems like a conservative lower bound for the minimum number of required training data points. The minimum test size based on Table 2 is 1K and is only valid for the NLP model with at least 20K samples of training data.

### 3.6. Factors affecting data set sizes

Many factors affect the amount of data needed to properly train and evaluate models. One factor is the prior class distribution in the evaluation data. In separate experiments, a minimum of 200 data points in the smaller class was found to be required. That value is consistent with numbers obtained in this study, in which test data set size should be about 1K, and for which the smallest class (rate of positive risk of depression) was roughly 20%. Another factor is the length of each sample, whether in time or words. The effect of test sample length on performance is explored in [35]. A third factor is the number of unique speakers. In [32] it was found that given a constant corpus budget of speaking time, it was better for acoustic depression detection to have shorter amounts of data from more speakers than more data from fewer speakers. Differences in speaker characteristics (demographics, medical conditions, dialects, and so on) all affect the levels of variability in the data. Data requirements are also affected by model quality: better models generally show lower variation for a fixed amount of test data. Additional factors affecting the amounts of data needed include audio quality and variability, speaking styles, task and metrics, and speech recognition performance, among others.

## 4. Conclusions and Future Directions

Spoken language technology offers promise for remote screening and monitoring of depression risk. Despite valuable progress in research studies, the use of small datasets makes it difficult to estimate performance robustness for clinical applications. Results from a fully crossed design of different train/test size combinations revealed several useful findings. First, test sizes below roughly 1K data points yielded too much variability. Second, training size results stabilized after about 2K samples. Values for minimum set sizes were given for different levels of tolerance in the amount of AUC variation. Third, NLP and acoustic models showed similar patterns with respect to train and test sizes. A difference was seen in terms of the amount of data needed for each model to begin to perform well. The NLP model was better overall, but the acoustic model was better able to cope with the small sample sizes (200, 500) that reflect the size of data used in almost all past studies. Fourth, results from testing on mismatched data showed a similar pattern of AUC variation to those in the matched data experiments, despite a lower overall performance for SP data.

Future work should focus on the need for larger train and test set sizes. Better understanding is needed of how various factors that influence the size requirements behave. Factors include but are not limited to: the nature of the data, class priors, model strength and pre-training, number of unique speakers, data lengths, and the metrics used in classification or regression. Overall, this study serves to provide a first set of reference results for performance robustness under a wide range of train and test data sizes. Results help estimate the level of benefit that can be achieved by using larger corpora in future studies of depression risk based on spoken language.

# 5. References

[1] S. Lewis, M. Freeman, M. van Ommeren, D. Chisholm, O. G. Siegl, D. Kestel, et al. "World Mental Health Report: Transforming Mental Health for All," *WHO*, Jun. 2022.

[2] B. R. Rittberg, "Major Depressive Disorder," in *The Medical Basis of Psychiatry, New York, NY: Springer New York*, 2016, pp. 79–90.

[3] D. M. Low, K. H. Bentley, and S. S. Ghosh, "Automated Assessment of Psychiatric Disorders Using Speech: A Systematic Review," *Laryngoscope Investig. Otolaryngol.* vol. 5, no. 1, pp. 96–116, 2020.

[4] S. H. Dumpala, S. Rempel, K. Dikaios, M. Sajjadian, R. Uher, and S. Oore, "Estimating Severity of Depression From Acoustic Features and Embeddings of Natural Speech," in *Proc. of ICASSP*, Jun. 2021, pp. 7278–7282.

[5] E. Villatoro-Tello, S. P. Dubagunta, J. Fritsch, G. Ramírez-de-la-Rosa, P. Motlicek, and M. Magimai-Doss, "Late Fusion of the Available Lexicon and Raw Waveform-Based Acoustic Modeling for Depression and Dementia Recognition," in *Proc. of INTERSPEECH*, Aug. 2021, pp. 1927–1931.

[6] S. Scherer, G. Stratou, J. Gratch, and L. P. Morency, "Investigating Voice Quality as a Speaker-Independent Indicator of Depression and PTSD," in *Proc. of INTERSPEECH*, 2013, pp. 847-851.

[7] M. Valstar et al., "AVEC 2016," in *Proc. of the 6th International Workshop on Audio/Visual Emotion Challenge*, Oct. 2016, pp. 3–10.

[8] F. Ringeval et al., "AVEC 2019 Workshop and Challenge: State-of-Mind, Detecting Depression with AI, and Cross-Cultural Affect Recognition," in *Proc. of the 9th International on Audio/Visual Emotion Challenge and Workshop*, 2019, pp. 3–12.

[9] M. Valstar et al., "AVEC 2014," in *Proc. of the 4th International Workshop on Audio/Visual Emotion Challenge*, 2014, pp. 3–10.

[10] M. P. M. Valstar, B. Schuller, K. Smith, F. Eyben, B. Jiang, S. Bilakhia, S. Schnieder, R. Cowie, "The Continuous Audio/Visual Emotion and Depression Recognition Challenge," in Proc. of 3rd ACM International Workshop on Audio/Visual Emotion Challenge, 2013, pp. 3–10.

[11] Shen Y, Yang H, Lin L. "Automatic Depression Detection: An Emotional Audio-Textual Corpus and a GRU/BiLSTM-based Model", *arXiv preprint arXiv:2202.08210*. 2022 Feb 15.

[12] M. Niu, K. Chen, Q. Chen and L. Yang, "HCAG: A Hierarchical Context-Aware Graph Attention Model for Depression Detection," in *Proc. of ICASSP*, 2021, pp. 4235-4239.

[13] D. Sztahó, K. Gábor, and T. Gábriel, "Deep Learning Solution for Pathological Voice Detection using LSTM-based Autoencoder Hybrid with Multi-Task Learning," in *Proc. of the 14th International Joint Conference on Biomedical Engineering Systems and Technologies*, 2021, pp. 135–141.

[14] N. Aloshban, A. Esposito, and A. Vinciarelli, "Language or Paralanguage, This is the Problem: Comparing Depressed and Non-Depressed Speakers Through the Analysis of Gated Multimodal Units," in *Proc. of INTERSPEECH*, Aug. 2021, pp. 2496–2500.

[15] L. Hansen, Y.-P. Zhang, D. Wolf, K. Sechidis, N. Ladegaard, and R. Fusaroli, "A Generalizable Speech Emotion Recognition Model Reveals Depression and Remission," *Acta Psychiatr. Scand.*, vol. 145, no. 2, pp. 186–199, Feb. 2022.

[16] J. C. Mundt, A. P. Vogel, D. E. Feltner, and W. R. Lenderking, "Vocal Acoustic Biomarkers of Depression Severity and Treatment Response," *Biol. Psychiatry*, vol. 72, no. 7, pp. 580–587, Oct. 2012.

[17] J. Dineley et al., "Remote Smartphone-Based Speech Collection: Acceptance and Barriers in Individuals with Major Depressive Disorder," in *Proc. of INTERSPEECH*, Aug. 2021, pp. 631–635.

[18] J. C. Mundt, P. J. Snyder, M. S. Cannizzaro, K. Chappie, and D. S. Geralts, "Voice Acoustic Measures of Depression Severity and Treatment Response Collected via Interactive Voice Response (IVR) Technology," *J. Neurolinguistics*, vol. 20, no. 1, pp. 50–64, Jan. 2007.

[19] N. Cummins, S. Scherer, J. Krajewski, S. Schnieder, J. Epps, and T. F. Quatieri, "A Review of Depression and Suicide Risk Assessment Using Speech Analysis," *Speech Commun.*, vol. 71, pp. 10–49, Jul. 2015.

[20] J. R. Williamson et al., "Detecting Depression using Vocal, Facial and Semantic Communication Cues," in *Proc. of the 6th International Workshop on Audio/Visual Emotion Challenge*, Oct. 2016, pp. 11–18.

[21] A. Pampouchidou et al., "Depression Assessment by Fusing High and Low Level Features from Audio, Video, and Text," in *Proc. of the 6th International Workshop on Audio/Visual Emotion Challenge*, Oct. 2016.

[22] Z. Zhao, Z. Bao, Z. Zhang, N. Cummins, H. Wang, and B. Schuller, "Hierarchical Attention Transfer Networks for Depression Assessment from Speech," in *Proc. of ICASSP*, May 2020, pp. 7159–7163.

[23] D. E. Nease and J. M. Malouin, "Depression Screening: a Practical Strategy," *J. of Family Practice*, vol. 52, no. 2, pp. 118– 126, 2003.

[24] L. Manea, S. Gilbody and D. Mcmillan, "Optimal Cut-off Score for Diagnosing Depression with the Patient Health Questionnaire (PHQ-9): a meta-analysis," *CMAJ*, vol. 184, 2011.

[25] K. Kroenke, T.W. Strine, R. Spitzer, J.B.W. Williams, J.T. Berry, and A.H. Mokdad, "The PHQ-8 as a Measure of Current Depression in the General Population," *J. Affect. Disord.*, vol. 114, no. 1–3, 2009.

[26] A. M. Dai and Q. V. Le, "Semi-supervised Sequence Learning," *Adv. Neural Inf. Process. Syst., Nov. 2015,* pp. 3079–3087.

[27] P. He, X. Liu, J. Gao, W. Chen, "Deberta: Decoding-enhanced Bert With Disentangled Attention", *arXiv preprint* arXiv:2006.03654, 2020.

[28] Y. Liu, et al., "Roberta: A Robustly Optimized Bert Pretraining Approach", *arXiv preprint* arXiv:1907.11692, 2019.

[29] Z. Lan, M. Chen, S. Goodman, K. Gimpel, P. Sharma, R. Soricut, "Albert: A Lite Bert for Self-supervised Learning of Language Representations", *arXiv preprint* arXiv:1909.11942, 2019.

[30] J. Devlin, M.-W. Chang, K. Lee, K. Toutanova, "Bert: Pre-training of Deep Bidirectional Transformers for Language Understanding", *arXiv preprint* arXiv:1810.04805, 2018.

[31] T. Rutowski, E. Shriberg, A. Harati, Y. Lu, P. Chlebek, and R. Oliveira, "Depression and Anxiety Prediction Using Deep Language Models and Transfer Learning," in *Proc. of 7th International Conference on Behavioral and Social Computing (BESC)*, Nov. 2020, pp. 1–6.

[32] A. Harati et al., "Generalization of Deep Acoustic and NLP Models for Large-Scale Depression Screening," Accepted for publication in Biomedical Sensing and Analysis: *Signal Processing in Medicine and Biology*, 2022.

[33] S. Lai, L. Xu, K. Liu, and J. Zhao, "Recurrent Convolutional Neural Networks for Text Classification," in *Proc. of National Conference on Artificial Intelligence,* 2015, pp. 2267– 2273.

[34] A. Harati, E. Shriberg, T. Rutowski, P. Chlebek, Y. Lu, and R. Oliveira, "Speech-Based Depression Prediction Using Encoder-Weight-Only Transfer Learning and a Large Corpus," in *Proc. of ICASSP*, Jun. 2021, pp. 7273–7277.

[35] T. Rutowski, A. Harati, Y. Lu, and E. Shriberg, "Optimizing Speech-Input Length for Speaker-Independent Depression Classification," in *Proc. of INTERSPEECH*, Sep. 2019, pp. 3023–3027.